\documentclass{article}
\usepackage{spconf,amsmath,epsfig,amssymb,amsthm}
\usepackage{graphicx}
\usepackage{cite}
\usepackage{multirow}
\usepackage{tabularx}
\usepackage{tabu}
\usepackage{subfigure}
\usepackage{paralist, tabularx}
\usepackage[english]{babel}
\DeclareMathOperator{\x}{x}
\DeclareMathOperator{\y}{y}
\DeclareMathOperator{\hh}{hh}
\DeclareMathOperator{\hv}{hv}
\DeclareMathOperator{\vv}{vv}

\title{Very high resolution Airborne PolSAR Image Classification using Convolutional Neural Networks}

\name{Minh-Tan Pham, S\'ebastien Lef\`evre}
\address{
Univ. Bretagne Sud - IRISA, UMR CNRS 6074, 56000 Vannes, France
}

\begin{document}
%
\maketitle
\begin{abstract}
In this work, we exploit convolutional neural networks (CNNs) for the classification of very high resolution (VHR) polarimetric SAR (PolSAR) data. Due to the significant appearance of heterogeneous textures within these data, not only polarimetric features but also structural tensors are exploited to feed CNN models. For deep networks, we use the SegNet model for semantic segmentation, which corresponds to pixelwise classification in remote sensing. Our experiments on the airborne F-SAR data show that for VHR PolSAR images, SegNet could provide high accuracy for the classification task; and introducing structural tensors together with polarimetric features as inputs could help the network to focus more on geometrical information to significantly improve the classification performance.
\end{abstract}


\section{Introduction}
The overwhelming development of deep learning based on convolutional neural networks (CNNs) in image processing and computer vision domains has provided useful tools and techniques to tackle various tasks in remote sensing. Research studies based on deep neural networks for the segmentation and classification of SAR/PolSAR data have been conducted within the past few years. Most literature studies propose to extract different polarimetric features from the scattering coherency matrix $T$ of PolSAR data and feed them into CNN models (see a review in \cite{parikh2019classification}). As a usual approach, the patch-based technique  divides a large PolSAR image into patches and applies a CNN to predict a single label for the center pixel \cite{zhou2016polarimetric, wang2018polarimetric, ma2019land}. In \cite{zhou2016polarimetric}, the authors proposed a 6-D real vector converted from the coherency matrix to encode polarimetric information and feed a 4-layer CNN which could provide performance similar to the state-of-the-art. Similarly, polarimetric features derived from Pauli representation as well as diagonal and off-diagonal components of $T$ were exploited in \cite{wang2018polarimetric, ma2019land} using this patch-based technique. However, as discussed in \cite{mohammadimanesh2019a}, such a patch-based approach could yield boundary artifacts and be less relevant than semantic segmentation networks using an encoder-decoder architecture such as FCNs (fully convolutional networks) \cite{long2015fully} or SegNet \cite{badrinarayanan2017segnet}. Therefore, this paper proposes to exploit rather an FCN or a SegNet model to perform pixelwise classification of PolSAR images based on their intensity and polarimetric information, and proved their effectiveness on common PolSAR data benchmarks used in the literature \cite{mohammadimanesh2019a}. 

The development of very high resolution (VHR) sensors offers PolSAR image data including not only the fully polarimetric characteristics but also the significant spatial information. Heterogeneous textures and structures become necessary to be taken into account for different PolSAR tasks including classification. To deal with high/very high spatial resolution remote sensing, texture information could not be neglected. In \cite{du2015random, pham2015,pham2018fusion}, the authors proposed to fuse polarimetric information with spatial features such as morphological profiles, textures or structural tensors when dealing with VHR PolSAR image classification. We continue this approach by extracting polarimetric and structural features of VHR PolSAR data, setting them as inputs of the SegNet model to perform pixelwise classification task. We argue that although CNN models are known to be able to extract and learn relevant features within an end-to-end approach, adding structural tensors as inputs together with polarimetric features might help the network to focus more on spatial information (particular crucial in such VHR PolSAR data), which finally improves the classification performance. 

The contributions of this paper are: 1/ We extract and incorporate structural tensors together with polarimetric features to tackle classification task of VHR PolSAR data; 2/ We consider the SegNet model for pixelwise classification with a disjoint split of training and test sets from the studied data; 3/ We provide qualitative and quantitative performance of our approach and show our improvement in terms of classification accuracy.

In the remainder of this paper, Section \ref{sec:data} provides the information about the VHR PolSAR data exploited in our work. Section \ref{sec:method} describes the proposed method including the extraction of polarimetric and structural features, as well as the use of SegNet model for classification task. We then show our experimental study on Section \ref{sec:exp} and then conclude the paper in Section \ref{sec:conclusion}.

\section{Dataset}
\label{sec:data}
For this study, we exploit a PolSAR image acquired in Kaufbeuren, Germany in 2010 by the VHR airborne F-SAR system (S-band) operated by the German Aerospace Center (DLR). The entire image consists of $8500\times 17152$ pixels with the pixel spacing in azimuth and range of 0.34 m and 0.3 m, respectively. A region of interest including $1800 \times 3000$ pixels was extracted to perform our experiments. The studied region is shown in Fig.~\ref{fig:data}-(a) using RGB Pauli color-coded representation (i.e. Red = $|$HH-VV$|$, Green = $|$HV$|$ and Blue = $|$HH+VV$|$). The thematic ground truth including 5 classes (i.e. tree, solar panel, grass pasture, building and road) was manually generated with the help of OpenStreetMap and Google Maps to provide the most relevant land-cover interpretation of the observed scene (see Fig. \ref{fig:data}-(b)).

\section{Methodology}
\label{sec:method}
\subsection{Polarimetric and structural features}
We first remind some particularities of PolSAR image data, starting with the complex polarimetric scattering vector in Pauli representation:
\begin{equation}
\begin{split}
k_p & = \frac{1}{\sqrt{2}}\big[S_{\hh}-S_{\vv},2S_{\hv},S_{\hh}+S_{\vv}\big]^T \\
& = [k_1,k_2, k_3]^T
\end{split}
\label{eq:kp}
\end{equation}

From $k_p$, the $3\times 3$ polarimetric coherency matrix $T$ and the total back-scattering power (SPAN) are calculated as: 
\begin{equation}
T = k_pk_p^T
\label{eq:t}
\end{equation}
\begin{equation}
Span = T_{11} + T_{22} + T_{33} = k_p^Tk_p
\label{eq:span}
\end{equation}

In the literature, two 6-D polarimetric feature vectors were derived from $T$ and SPAN and used as inputs of the CNNs in \cite{ma2019land} and \cite{zhou2016polarimetric}, respectively as follows:
\begin{equation}
vec(T)= \big[ \thickspace T_{11}, T_{22}, T_{33}, |T_{12}|, |T_{13}|, |T_{23}| \thickspace \big]
\label{eq:6D1}
\end{equation}
\begin{equation}
\begin{split}
vec(Span,T) = \big[& \log_{10}(Span), \thickspace \frac{T_{22}}{Span}, \thickspace \frac{T_{33}}{Span}, \\
 & \frac{|T_{12}|}{\sqrt{T_{11}T_{22}}}, \frac{|T_{13}|}{\sqrt{T_{11}T_{33}}}, \frac{|T_{23}|}{\sqrt{T_{22}T_{33}}} \thickspace \big]
\end{split}
\label{eq:6D}
\end{equation}

Unlike most literature studies which exploit only polarimetric features from $k_p$, $T$ or SPAN as above, we propose to extract the structural tensors based on the principle of Di Zenzo gradients \cite{di1986a} as proposed in \cite{pham2015,pham2018fusion}:
\begin{subequations}
\begin{equation}
  J_{\x\x} = \sum_{i=1}^3 \left( \frac{\partial |k_i|}{\partial x} \right)^2
  \label{subeq:Jxx}
\end{equation}    
\begin{equation}
  J_{\x\y} = \sum_{i=1}^3 \left( \frac{\partial |k_i|}{\partial x} \right)\left( \frac{\partial |k_i|}{\partial y} \right)
\end{equation}
\begin{equation}
  J_{\y\y} = \sum_{i=1}^3 \left( \frac{\partial |k_i|}{\partial y}\right)^2
  \label{subeq:Jyy}
\end{equation}
\label{eq:tensor}
\end{subequations}
where $k_i, i=1,\ldots,3$ are the 3 element of $k_p$ \eqref{eq:kp} and the computation of horizontal and vertical derivatives is adapted for SAR images using mean ratio operator:
\begin{subequations}
\begin{equation}
  \frac{\partial v}{\partial x} = 1- \min \bigg\{ \frac{v(x+1,y)}{v(x-1,y)},\frac{v(x-1,y)}{v(x+1,y)} \bigg\}
\end{equation}    
\begin{equation}
 \frac{\partial v}{\partial y} = 1- \min \bigg\{ \frac{v(x,y+1)}{v(x,y-1)},\frac{v(x,y-1)}{v(x,y+1)} \bigg\}
\end{equation}
\label{eq:dev}
\end{subequations}

\begin{figure}[t!]
      \centering
        \includegraphics[width=\linewidth]{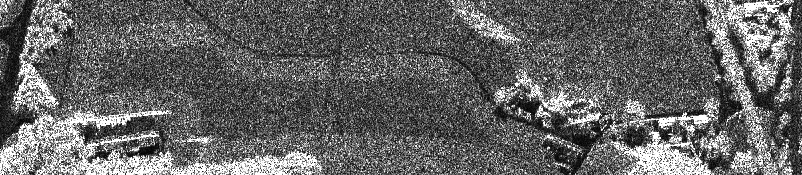} \\
        \footnotesize{(a) Polarimetric band $S_{hv}$} \\
        \vspace{1mm}
        \includegraphics[width=\linewidth]{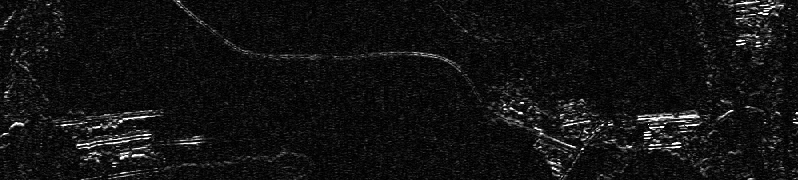}\\
        \footnotesize{(b) Tensor $J_{\x\x}$} \\
        \vspace{1mm}       
        \includegraphics[width=\linewidth]{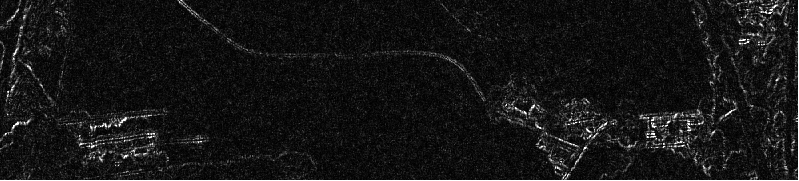}\\
        \footnotesize{(c) Jyy} \\
  \caption{Illustration of the tensors $J_{\x\x}$ and $J_{\y\y}$ which help to characterize structural and textural information not clearly present with polarimetric bands}
  \label{fig:illus}
  \end{figure}
To replace the 6-D polarimetric features in \eqref{eq:6D1} and \eqref{eq:6D}, we propose to exploit a new 6-D feature vector which incorporates polarimetric and structural features: 
\begin{equation}
vec(k_p,J) = \big[|k_1|, |k_2|, |k_3|, J_{\x\x} ,J_{\x\y}, J_{\y\y} \thickspace \big]
\label{eq:our}
\end{equation}

\subsection{SegNet for pixelwise classification}
As discussed in our introduction, we consider in this work the SegNet model \cite{badrinarayanan2017segnet} which is one of the state-of-the-art semantic segmentation networks in the computer vision domain. We note that the authors in \cite{mohammadimanesh2019a} have investigated both FCNs and SegNet models and proved their equivalent capacity to perform pixelwise classification task (i.e. semantic segmentation in computer vision) of PolSAR data. Without lack of generality, SegNet is considered in our study but any other semantic segmentation CNNs could also be used to perform the similar task. Beside the mentioned work in \cite{mohammadimanesh2019a}, SegNet has also proved its effectiveness to tackle pixelwise classification in optical remote sensing such as multispectral images with visible (RGB) and infrared bands in \cite{audebert2018beyond}; as well as in Lidar rasters as proposed in \cite{guiotte2020}. 

\begin{figure}[h!]
      \centering
        \includegraphics[width=\linewidth]{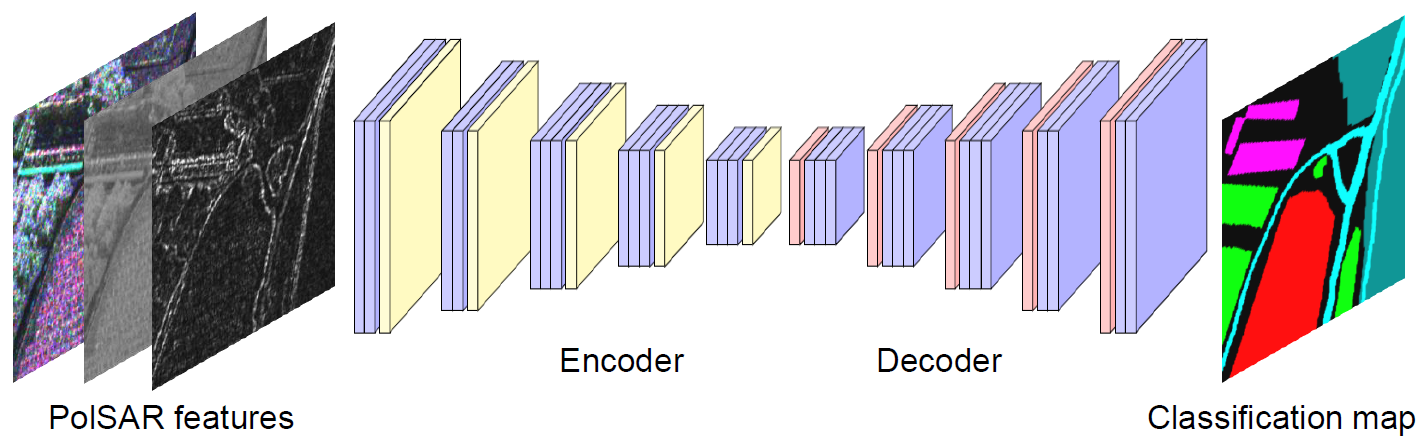}
  \caption{SegNet architecture for pixelwise classification of VHR PolSAR data}
  \label{fig:segnet}
  \end{figure}
  
As a short reminder, this network relies on an encoder-decoder architecture based on the convolutional layers of the VGG-16 network (see Fig. \ref{fig:segnet}), followed by batch normalization, rectified linear unit (ReLU) and then pooling and unpooling layers (w.r.t the encoder and decoder parts, respectively) \cite{badrinarayanan2017segnet}. For more details, readers are invited to the original paper. In our work, we generally set parameters as default for training and prediction phases which will be provided in the next section.

\section{Experimental study}
\label{sec:exp}

\begin{figure*}
      \centering
      \begin{tabular}{cc}
      \includegraphics[width=0.48\linewidth]{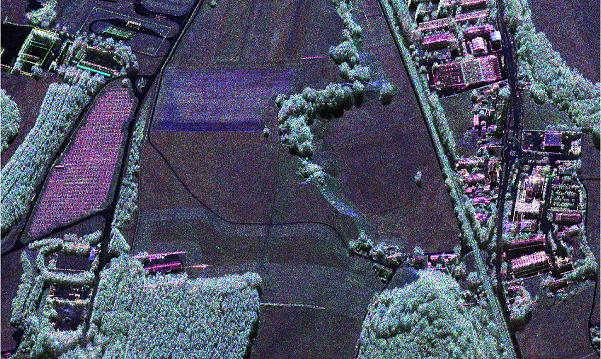} & \includegraphics[width=0.47\linewidth]{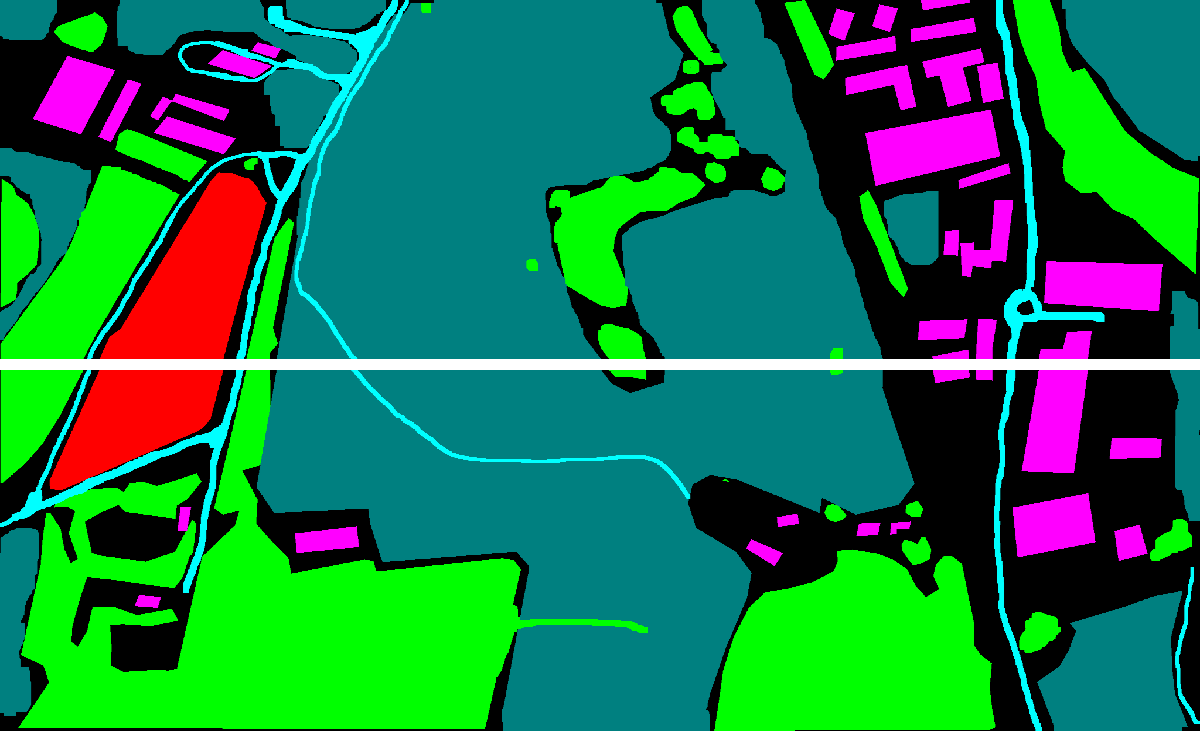} \\
      \footnotesize{(a) PolSAR image  ($1800\times 3000$ pixels) with Pauli color code}  & \footnotesize{(b) Horizontal disjoint split of training set (bottom) and test set (top)} \\
      
       \includegraphics[width=0.48\linewidth]{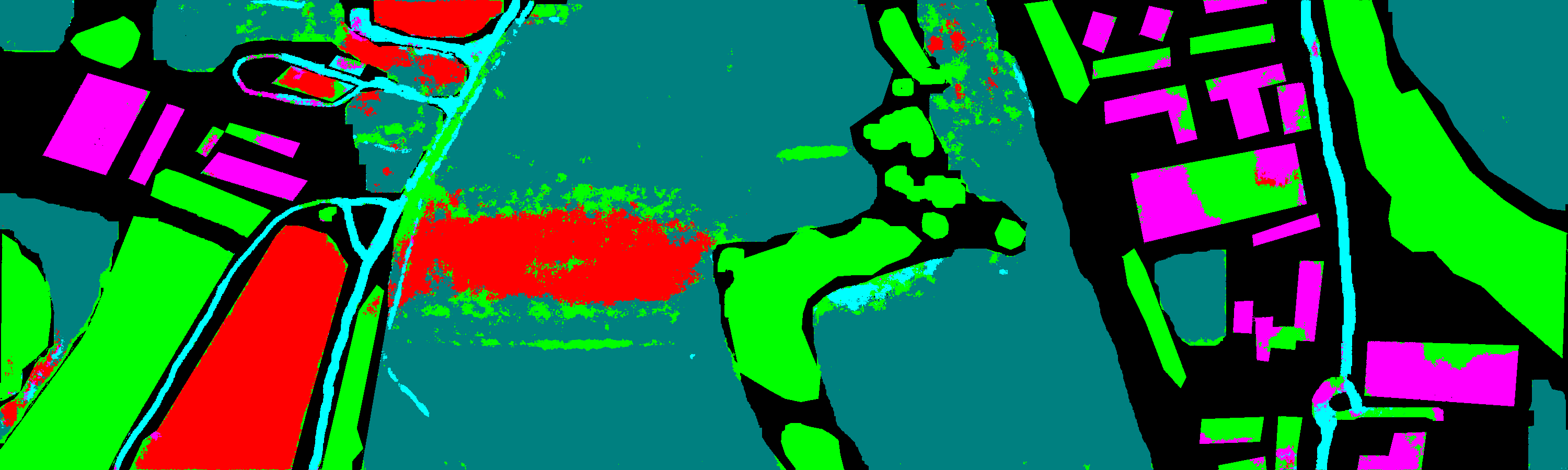} & \includegraphics[width=0.48\linewidth]{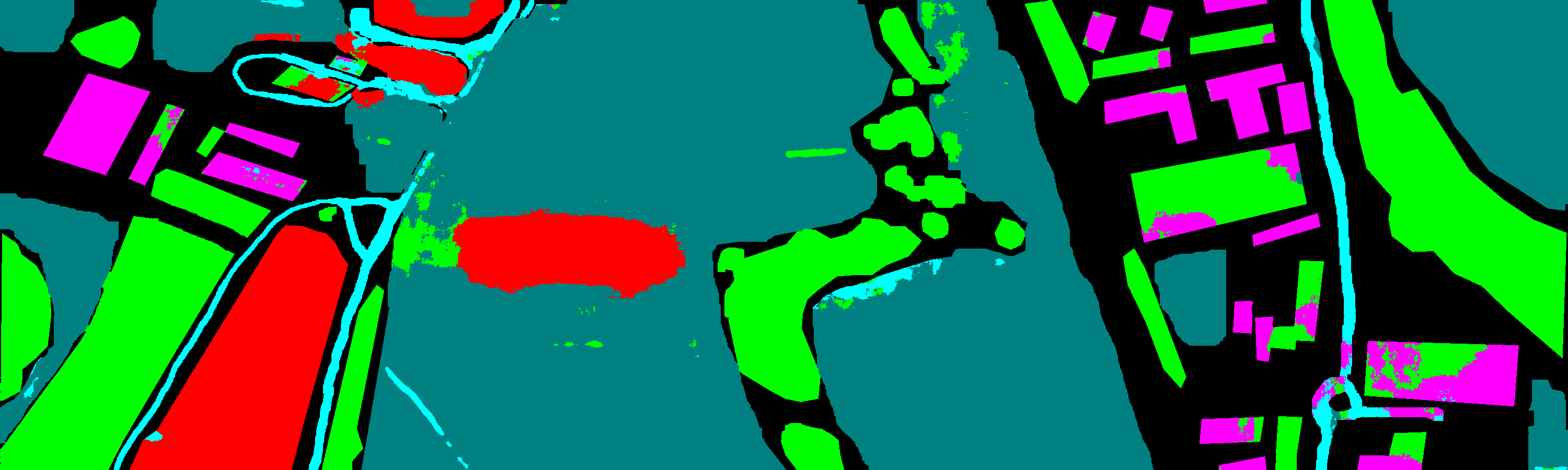}\\
       \footnotesize{(c) SPAN}  & \footnotesize{(d) Pauli} \\
       
       \includegraphics[width=0.48\linewidth]{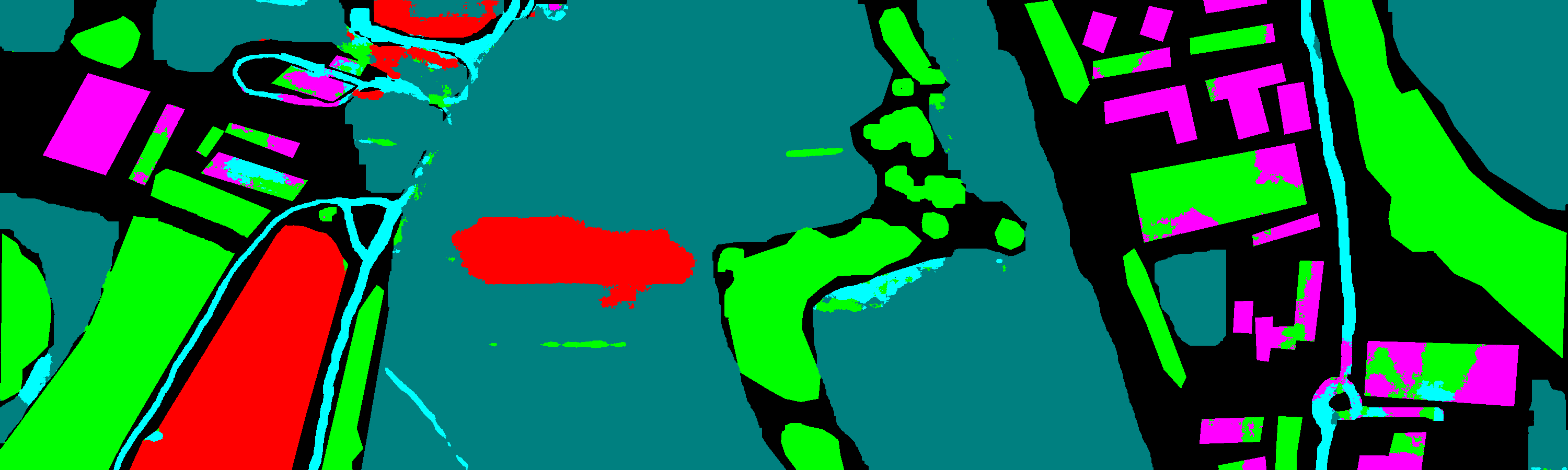} & \includegraphics[width=0.48\linewidth]{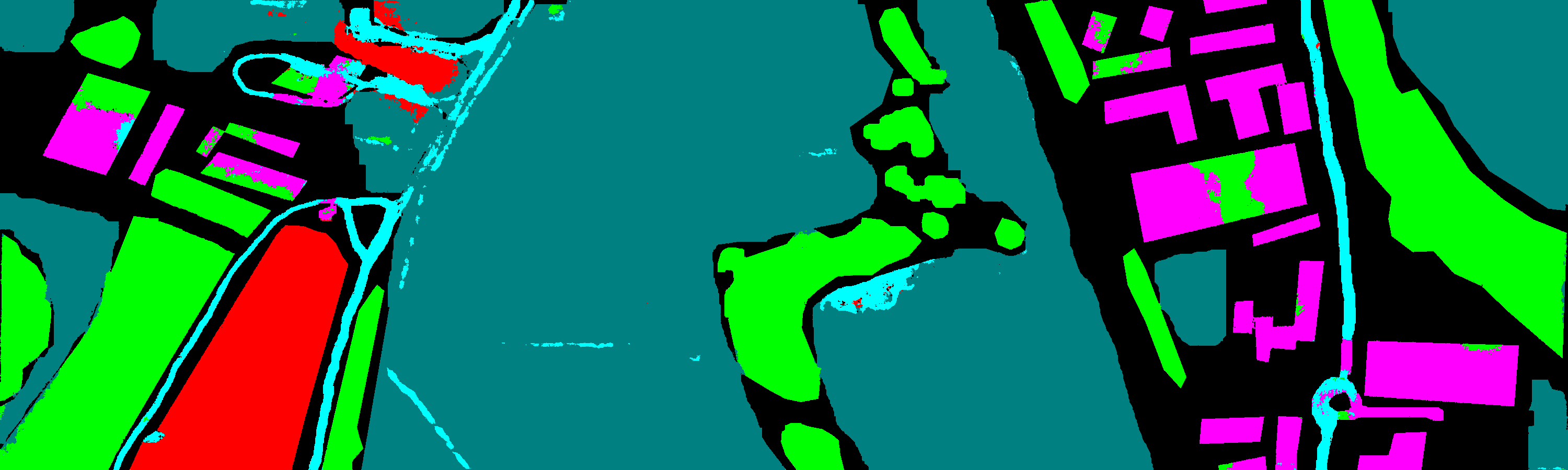}\\
       \footnotesize{(e) 6-D vector}  & \footnotesize{(f) Pauli+Tensor} \\
      \end{tabular}    
  \caption{Image data, training/test disjoint split, classification maps obtained by different approaches}
  \label{fig:data}
  \end{figure*}

\begin{table*}[!ht]
	\centering 
	\resizebox{0.995\textwidth}{!}{%
	{\renewcommand{\arraystretch}{1.1}
	\begin{tabu}{l| c |c c c c c |c c c}
	\hline
	\multirow{2}{*}{\textbf{Features}} & \multirow{2}{*}{\textbf{Channel}} & \multicolumn{5}{c|}{\textbf{Per-class accuracy (F1-score $\%$)}} & \multicolumn{3}{c}{\textbf{Overall performance}} \\
	\cline{3-10}
	 & & Tree & Solar Panel & Building & Grass & Road & AA ($\%$) & OA ($\%$) & \emph{$\kappa$} \\
	\hline
	\hline
	SPAN \eqref{eq:span} & 1 & 84.67 & 62.42  & 80.45 & 91.09 & 79.58 & 79.64 & 85.92 & 0.7721\\
	\hline
	Pauli $k_p$ \eqref{eq:kp} \cite{wang2018polarimetric} & 3 & 88.53 & 70.42  & 73.80 & 94.56 & 79.93 & 81.45 & 89.20 & 0.8187\\
	\hline
	6-D vector \eqref{eq:6D} \cite{zhou2016polarimetric} & 6 & 94.27 & 86.00 & 77.96 & 96.28 & {\bf 80.48} & 87.00 & 93.09 & 0.8785\\
	\hline
	Tensor $J$ \eqref{eq:tensor} & 3 & 90.13 & 88.76 & 84.08 & 97.53 & 71.81 & 86.46 & 93.30 & 0.8834\\
	\hline
	{\bf Pauli $k_p$+Tensor $J$} \eqref{eq:our} & 6 & {\bf 95.32} & {\bf 91.63} & {\bf 88.87} & {\bf 97.90} & 76.78 & {\bf 90.10} & {\bf 95.29} & {\bf 0.9184}\\
	\hline
	\end{tabu}
	}
	}
	\caption{Comparison of classification performance yielded by different approaches}
	\label{tab:results}
\end{table*}

\subsection{Setup}
The studied region has been divided into a training set and a test set with an horizontal split as shown on Fig.\ref{fig:data}-(b). This disjoint splitting technique allows us to train and test the network on separate pixel samples. 
We used the code\footnote{https://github.com/nshaud/DeepNetsForEO} from \cite{audebert2018beyond}
to perform all the experiments with parameter setting as default (learning rate $0.01$ with momentum 0.9 and weight decay $5\times 10^{-4}$) for a fair comparison. We note that the number of input channels of our SegNet model varies according to the features used as input (as shown later in Table \ref{tab:results}). The network was trained from scratch and during training phase, we randomly extracted $256\times 256$ patches from the training set to feed the network. Batch size was set to 16 and we reported the best performance within the first 20 epochs for all experiments.

\subsection{Results}
Fig. \ref{fig:data} and Table \ref{tab:results} show the qualitative and quantitative classification results on the studied VHR PolSAR data using different features as input of the SegNet model including the SPAN, the Pauli representation $k_p$ as \cite{wang2018polarimetric}, 6-D feature vector $vec(Span,T)$ proposed by \cite{zhou2016polarimetric}, the structural tensor $J$ and the proposed combination of $k_p$ and $J$. As observed from the figure, all approaches using only polarimetric features from the SPAN, $k_p$ and $T$ usually mis-classify classes that have similar color information such as: solar panel vs building, solar panel vs a part of grass pasture (which contains more soil and less grass at the center of the test image). These remarks can be observed from Fig. \ref{fig:data}-(c,d,e) where a large region of grass pasture (dark green) was mis-classified as solar panel (red). By introducing structural tensors, we have overcome this problem thanks to the significant dissimilarity between homogeneous texture of grass pasture vs the structured patterns within the solar panel, as observed in Fig. \ref{fig:data}-(f). In addition, the building class (magenta) has been better classified with the integration of structural information. 

The quantitative improvement can now be observed from Table \ref{tab:results}. Here, we report the classification accuracy for each thematic class \emph{(F1-score)}, the \emph{average accuracy (AA)}, the \emph{overall accuracy (OA)} and the \emph{kappa coefficient ($\kappa$)}. Our first remark is that SegNet is a relevant deep neural network to tackle pixelwise classification of VHR PolSAR data. By only using the SPAN information, it could yield an OA of 85.92\% and $\kappa$ of 0.77. Then, better polarimetric features from $k_p$ or $T$ could provide a $k$ equal to 0.82 and 0.88, respectively.  By adding structural tensors, we have improved the classification performance with an OA of 95.29\% and $k$ of 0.92. Another remark is that only using the tensor $J$ as input of SegNet could also provide equivalent performance than the 6-D vector encoding polarimetric features proposed from the literature \cite{zhou2016polarimetric}. This emphasizes the significance of spatial information (beside polarimetric characteristics) in particular when dealing with PolSAR images of very high spatial resolution.

\section{Conclusion}
\label{sec:conclusion}
Our study has found that semantic segmentation networks like SegNet could provide promising classification results on VHR polarimetric SAR data. To better guiding the network to focus on highly geometric and structural information from these data, we have proposed to further extract and incorporate structural gradient tensors with polarimetric features at inputs. Classification performance has been significantly improved by this manual guidance technique. As complex CNNs have been recently proposed to directly work on complex-valued PolSAR data, our future work is to exploit these novel networks with a remaining focus on geometric and textural information.

\section{Acknowledgement}
The authors would like to thank Dr. Andreas Reigber from the German Aerospace Center for providing the F-SAR data and Dr. Nicolas Audebert from the Conservatoire National des Arts et M\'etiers for sharing his code.


\end{document}